\documentclass[twocolumn]{article}

\usepackage[width=17cm,height=22cm]{geometry}
\usepackage[english]{babel}
\usepackage[utf8]{inputenc}
\usepackage{fancyvrb}
\usepackage{authblk}
\usepackage{graphicx}
\usepackage{hyperref}
\usepackage{amsmath, amssymb}

\usepackage{xcolor}
\usepackage{placeins}
\usepackage{enumitem}

\usepackage[utf8]{inputenc}
\usepackage{amssymb,amsmath,array}
\usepackage{xcolor}
\usepackage{subcaption}
\usepackage{placeins}

\usepackage[numbers]{natbib}

\begin{document}
%style file for ESANN manuscripts
%\title{Pixel-wise Conditioning of Generative Adversarial Networks}

\title{Pixel-wise Conditioning of Generative Adversarial Networks}

%***********************************************************************
% AUTHORS INFORMATION AREA
%***********************************************************************

\author{Cyprien Ruffino$^1$, Romain Hérault$^1$, Eric Laloy$^2$, Gilles Gasso$^1$
%
% Optional short acknowledgment: remove next line if non-needed
\thanks{This research was supported by the CNRS PEPS I3A  REGGAN project and the ANR-16-CE23-0006 grant \emph{Deep in France}}
%
% DO NOT MODIFY THE FOLLOWING '\vspace' ARGUMENT
\vspace{.3cm}\\
%
% Addresses and institutions (remove "1- " in case of a single institution)
1- Normandie Univ, UNIROUEN, UNIHAVRE, INSA Rouen, LITIS\\ 76~000 Rouen, France
%
% Remove the next three lines in case of a single institution
\vspace{.1cm}\\
2- Belgian Nuclear Research, Institute Environment, Health and Safety,\\ Boeretang 200 - BE-2400 Mol, Belgium
}

\maketitle

		\begin{figure*}[!]
		    \label{fig:mnist_generation}
		    \centering
		    \begin{subfigure}[t]{0.25\textwidth}
		        \centering
		        \includegraphics[scale=1.95]{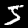}
		        \caption{Original \\ Image}
                \label{fig:digit}
            \end{subfigure}\begin{subfigure}[t]{0.25\textwidth}
		        \centering
		        \includegraphics[scale=8]{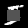}
		        \caption{Regular\\Inpainting}
                \label{fig:inpainting}
		    \end{subfigure}\begin{subfigure}[t]{0.25\textwidth}
		        \centering
		        \includegraphics[scale=8]{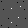}
		        \caption{Scatter\\Inpainting}
                \label{fig:pixelwise_gen}
		    \end{subfigure}\begin{subfigure}[t]{0.24\textwidth}
		        \centering
		        \includegraphics[scale=1.95]{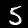}
		        \caption{Generated\\Image}
                \label{fig:generated}
		    \end{subfigure}
		    \caption{Difference between regular inpainting (b) and the problem undertaken in this work (c). The 
		    image obtained with our framework is shown in (d).}
		    \label{fig:image_completion}
		\end{figure*}
    
\begin{abstract}
   Generative Adversarial Networks (GANs) have proven successful for unsupervised image generation. Several works extended GANs to image inpainting by conditioning the generation with parts of the image one wants to reconstruct. However, these methods have limitations in settings where only a small subset of the image pixels is known beforehand. In this paper, we study the effectiveness of conditioning GANs by adding an explicit regularization term to enforce pixel-wise conditions when very few pixel values are provided. In addition, we also investigate the influence of this regularization term on the quality of the generated images and the satisfaction of the conditions. Conducted experiments on MNIST and FashionMNIST show evidence that this regularization term allows for controlling the trade-off between quality of the generated images and constraint satisfaction.
\end{abstract}

\section{Introduction}

	%Generative Adversarial Networks \cite{Goodfellow2014} have been the state of the art in image generation for the past few years, being able to produce realistic images with high resolution \cite{Brock2018}. Conditional Generative Adversarial Networks (CGAN) \cite{mirza2014} are a variant of GANs that can take into account additional information during the generation process. Since various types of information that can used, CGANs enable a variety of conditioned generation, such as class-conditioned image generation \cite{mirza2014}, or image-to-image translation \cite{Isola2017}.
	
	%At the opposite, our goal is to constrain GANs to generate images with predefined values on a subset of pixels. Such as, this resembles an inpainting task, which consists in reconstructing missing or altered parts of an image. However, inpainting tasks usually consists in "filling holes" in images. The "holes" are customary of small size, spatially located in the image and are structured parts of the image. 
	
	In this work we consider an extreme setting of inpainting task: we assume that only a few pixels, less than a percent of the considered image size, are known and that these pixels are randomly scattered across the image (see Fig.\ref{fig:pixelwise_gen}). This raises the challenge of how to take advantage of this scarce and unstructured a priori information to generate high quality images. Besides methodological novelty, a method that can tackle this problem would find applications to GAN-based geostatistical simulation and inversion in the geosciences \cite{laloy2018}.
	
    More specifically, this paper proposes an extension of the Conditional Generative Adversarial Network (CGAN) \cite{mirza2014} framework to learn the distribution of the training images given the constraints (the known pixels). To make the generated images honoring the prescribed pixel values, we use a regularization term measuring the distance between the real constraints and their generated counterparts. 
    Thereon we derive a learning scheme and analyze the influence of the used regularization term on both the quality of the generated images and the fulfillment of the constraints. By experimenting with a wide range of values for the additional hyper-parameter introduced by the regularization term, we show  for the MNIST~\cite{Lecun1998} and FashionMNIST~\cite{Xiao2017} datasets that our approach is effective and allows for controlling the trade-off between the quality of the generated samples and the satisfaction of the constraints.

\section{Related works}

%    In this section, we introduce the Generative Adversarial Networks\cite{Goodfellow2014} (GANs) framework and its conditional variant. We also present some of the existing methods that enable using GANs for image completion.
    
    %\subsection{Generative Adversarial Networks}
    %\bigskip
		Generative Adversarial Networks \cite{Goodfellow2014} basically consist of an algorithm for training generative models in an unsupervised way. It relies on a game between a generator, $G$, and a discriminator network, $D$, in which $G$ learns to produce new data with similar spatial characteristics/patterns as in the true data while $D$ learns to distinguish real examples from generated ones. Training GANs is equivalent to finding a Nash equilibrium to the following mini-max game:
		\vspace{-0.5em}
		\begin{align}
			&\min_G \max_D L(D, G) = \nonumber \\ 
			&\mathop{\mathbb{E}}_{x\sim P_r} \Big[\log (D(x))\Big] + \mathop{\mathbb{E}}_{z\sim P_z} \Big[\log (1-D(G(z)))\Big]
		\end{align}

		\noindent where $P_z$ is a known distribution, usually normal or uniform, in which latent variables are drawn, and $P_r$ is the distribution of the real samples.
		%
		% Je crois vraiment qu'on en a pas besoin de le dire ->
		%However, this objective has been shown to be particularly unstable, so in practice a non-saturating version of this cost is used , as recommended in \cite{Goodfellow2014}.

    %\subsection{Image completion with Generative Adversarial Networks}
    %\bigskip
        %As previously mentioned, the task studied in this paper can be seen as a special case of image completion in which less than a percent of the pixels are available and are randomly scattered across the image.
     % 
        %However, several previous works were done on image completion with GANs.
        
        Yeh et al. \cite{Yeh2017} introduced an inpainting method  which consists of taking a pre-trained generator and exploring its latent space $\mathcal{Z}$ via gradient descent, to find a latent vector, $z$, which induces an image close to the altered one while its quality remains close to the real samples. This method was applied by Mosser et al. \cite{Mosser} for 3D image completion with few constraints. However, the location of the constraints in their approach was fixed, instead of randomly scattered.
        
        Some other approaches rely on Conditional Generative Adversarial Networks~(CGAN) \cite{mirza2014}. This is a variant of GANs in which additional information,~$c$, is given to both the generator and the discriminator as an input (see Fig.\ref{fig:cgan}). The optimization problem becomes:
        %This allows the conditioning of the generator by using a conditional version of the GAN cost function :
		\vspace{-0.5em}
		\begin{align}
		&\min_G \max_D L(D, G) =\nonumber\\
		&\mathop{\mathbb{E}}_{\substack{x\sim P_{r}\\ \tilde{c}\sim P_{c|x} }} \Big[\log(D(x, \tilde{c}))\Big]
		\small{+} \mathop{\mathbb{E}}_{\substack{z\sim P_z\\c\sim P_{c}}} \Big[ \log(1\small{-}D(G(z,c),c))\Big]
		\end{align}
		
        In it seminal version \cite{mirza2014}, CGANs are used for class-conditioned image generation by giving the labels of the images to the networks. However, several kind of conditioning data can be used even a full image to do image-to-image translation \cite{Isola2017} or image inpainting \cite{Yu2018,Demir2018}.
        %This is the approach used in the aforementioned methods, as the authors provided the whole altered image to the CGAN to complete the missing parts.

        %These methods perform well to "fill holes" in images but are defeated in our particular setting where only few parts of the images are known. \remarqueCR{Non, pas forcément. On le voit d'ailleurs sur la figure \ref{fig:mses}, pour $\lambda=0$} 
       
\section{Proposed approach}

    In this work, we retain the CGAN approach and add a reconstruction loss term to further enforce the prescribed pixel values (usually less than a percent of the image). With this setup, the generator can be used to generate images from constraints unseen during the training.
    
    %\begin{figure}[t]
    %\centering
	%    \includegraphics[scale=0.3]{approach.png}
	%    \caption{Overview of our approach.
	%    %We use the CGAN setting, to which we add a $L2$ regularization cost between the constraints and the constrained pixels of the generated image
	%    }
	%    \label{fig:approach}
	%\end{figure}
        
        \begin{figure}[t]
		    \centering
		    %\begin{subfigure}[b]{0.45\textwidth}
		    %    \includegraphics[width=\textwidth]{gan.pdf}
		    %    \caption{GAN}
            %    \label{fig:gan}
		    %\end{subfigure}
		    \begin{subfigure}{0.45\textwidth}
		        \includegraphics[width=\textwidth]{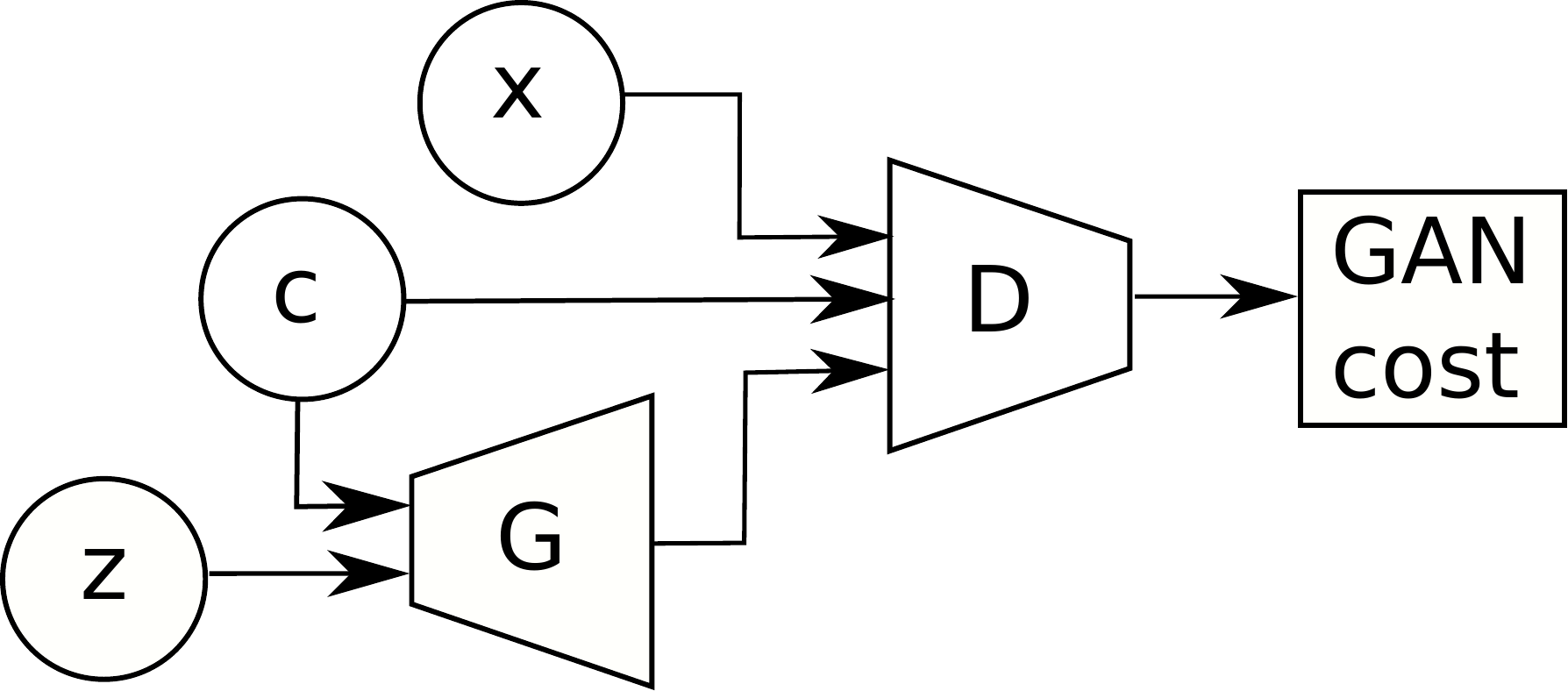}
		        \caption{CGAN}
                \label{fig:cgan}
		    \end{subfigure}

		    \begin{subfigure}{0.45\textwidth}
		        \includegraphics[width=\textwidth]{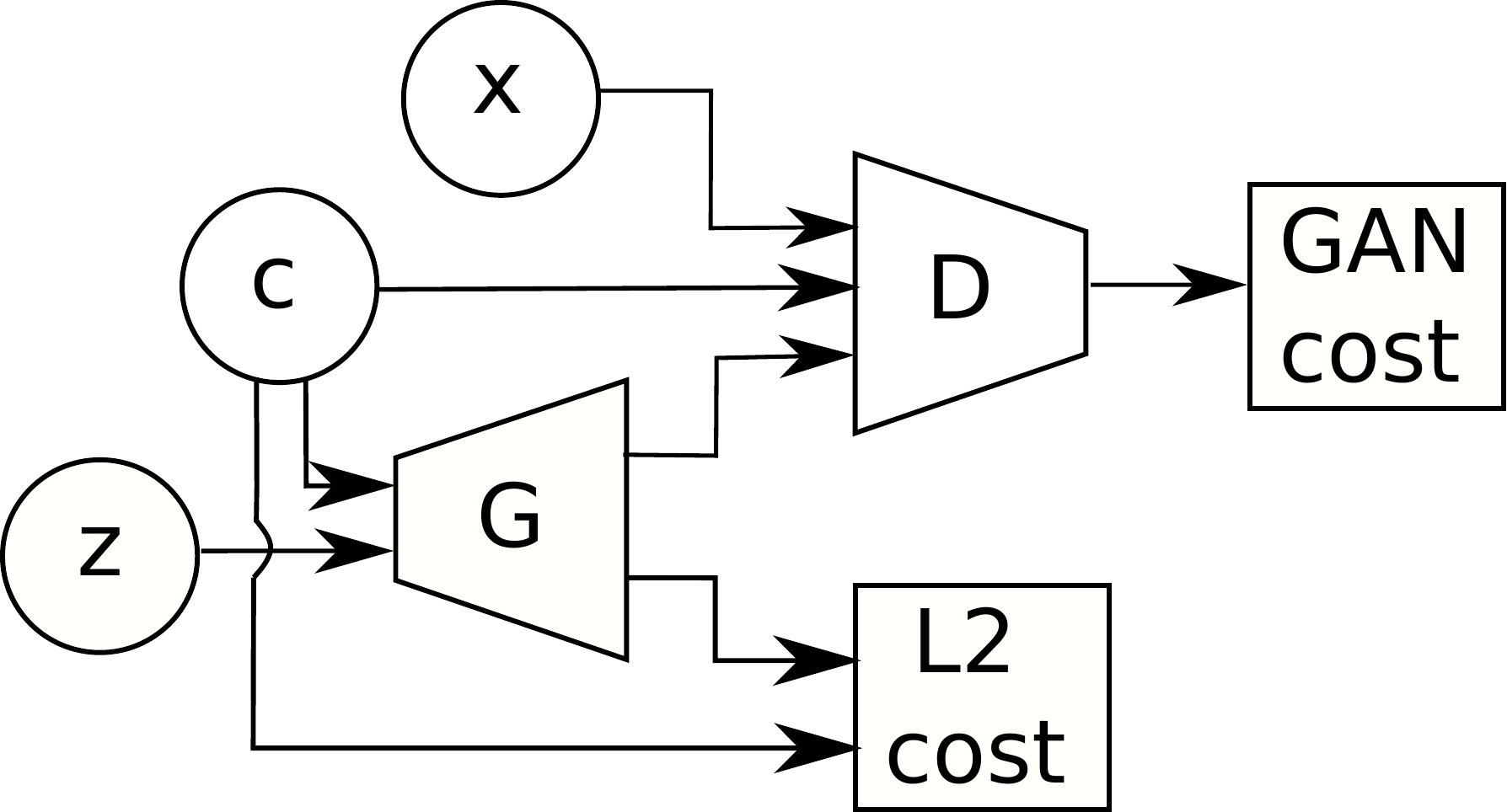}
		        \caption{Our approach}
                \label{fig:ourapproach}
		    \end{subfigure}
		    \caption{Different GAN Setups}
		    \label{fig:gansetup}
		\end{figure}

    	\begin{figure*}
		    \centering
		    \begin{subfigure}[t]{0.25\textwidth}
		        \centering
		        \includegraphics[scale=1.5]{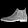}
		        \caption{Original\\Image}
                \label{fig:original_shoe}
            \end{subfigure}\begin{subfigure}[t]{0.25\textwidth}
		        \centering
		        \includegraphics[scale=1.5]{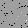}
		        \caption{Constraints}
                \label{fig:constraints}
		    \end{subfigure}\begin{subfigure}[t]{0.25\textwidth}
		        \centering
		        \includegraphics[scale=1.5]{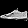}
		        \caption{Generated\\Image}
                \label{fig:pixelwise}
		     \end{subfigure}\begin{subfigure}[t]{0.24\textwidth}
		        \centering
		        \includegraphics[scale=1.5]{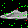}
		        \caption{Satisfied\\Consts.}
                \label{fig:generated}
		    \end{subfigure}
		    \caption{Generation of a sample during training. We first sample an image from a training set (a) and we sample the constraints from it. Then our GAN generates a sample (c). The constraints with squared error smaller than $\epsilon=0.1$ are deemed satisfied and shown by green pixels in (d) while the red pixels are unsatisfied.}
		    \label{fig:image_completion}
		\end{figure*}

    Given a learning set of images $X \in [-1, 1]^{P\times P}$ drawn from an unknown distribution $P_r$ and a sparse matrix  $C \in  [-1, 1]^{P\times P}$ as the given constrained pixels, the problem we focused on consists in finding a generative model $G$ with input $z\sim P_z$, a random vector sampled from a known distribution, and constrained pixel values $\tilde{C} \in  [-1, 1]^{P\times P}$ that could generate an image satisfying the constraints while likely following the distribution $P_r$.
    Enforcing the constraints in the CGAN framework leads to the following problem:
	\begin{align}
	\label{eq:formulation_our_primary_CGAN}
		&\min_G \max_D L(D,G) {\small=}\nonumber\\
		&\mathop{\mathbb{E}}_{\substack{X\sim P_{r}\\ \tilde{C}{\small\sim} P_{C|X}}} \Big[\log(D(X, \tilde{C}))\Big]
		{\small+} \mathop{\mathbb{E}}_{\substack{z{\small\sim} P_z\\C{\small\sim} P_{C}}} \Big[ \log(1{\small -}D(G(z , C), C))\Big]\nonumber \\
		&\text{s.c. } C = M(C) \odot G(z,C))
		%\quad \, \forall\, z,C \sim P_z,P_{C}  \nonumber
	\end{align}
        
    \noindent where $\odot$ is the Hadamard (or point-wise) product and $M(C)$ is a corresponding masking matrix. $M(C)$ is a sparse matrix with entries equal to one at constrained pixels location. As the equality constraint in  \ref{eq:formulation_our_primary_CGAN} is hard to enforce during training, we rather investigate a relaxed version of the problem. Indeed, we minimize the $L_2$ norm between the constrained pixels and the generated values (see Fig.\ref{fig:ourapproach}).
	The objective function, with $\lambda \geq 0$ a regularization parameter, becomes:
	\begin{align}
		&L(D,G) = \mathop{\mathbb{E}}_{\substack{X\sim P_{r}\\\tilde{C}\sim P_{C|X}}} \Big[\log(D(X, \tilde{C}))\Big]\nonumber\\
        &+\mathop{\mathbb{E}}_{\substack{z\sim P_z\\C\sim P_{C}}} \Big[\log(1-D(G(z, C), C))\Big] \nonumber \\
		& +\mathop{\mathbb{E}}_{\substack{z'\sim P_z\\'C\sim P_{C}}} \Big[\lambda\left\|C' - M(C') \odot G(z,C')\right\|_2^2\Big]
	\end{align}
	 
     %The hyperparameter $\lambda$ allows the explicit control of the tradeoff between the quality of the generated samples and the enforcing of the conditions, while keeping the GAN training algorithm unchanged. \remarqueCR{Ca, c'est ce qu'on conclut après les expériences, non ?}
     
     \begin{figure*}[!]
            \centering
            \includegraphics[trim=0 0 0 40, clip,scale=0.36]{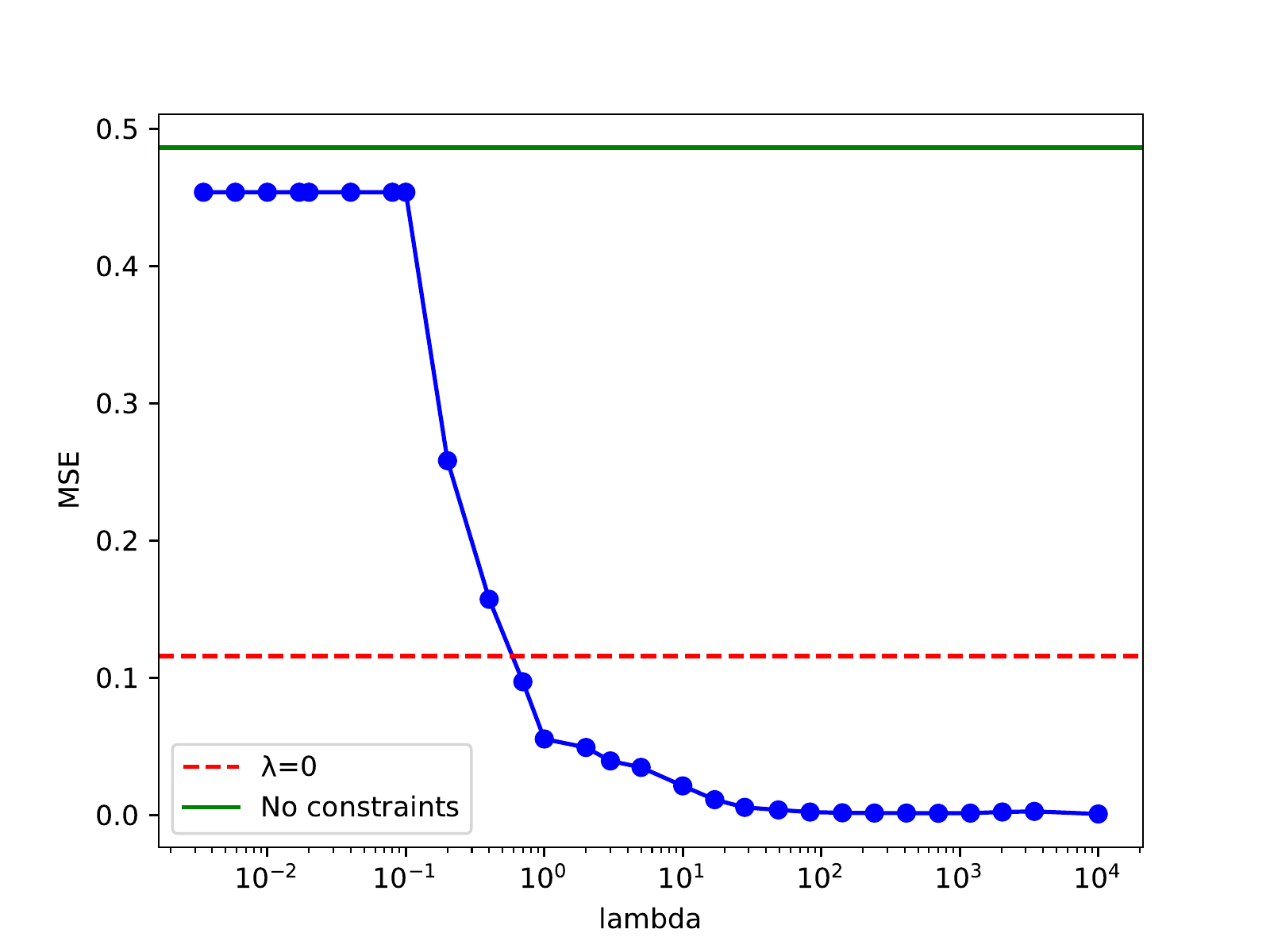}\includegraphics[trim=0 0 0 40, clip,scale=0.36]{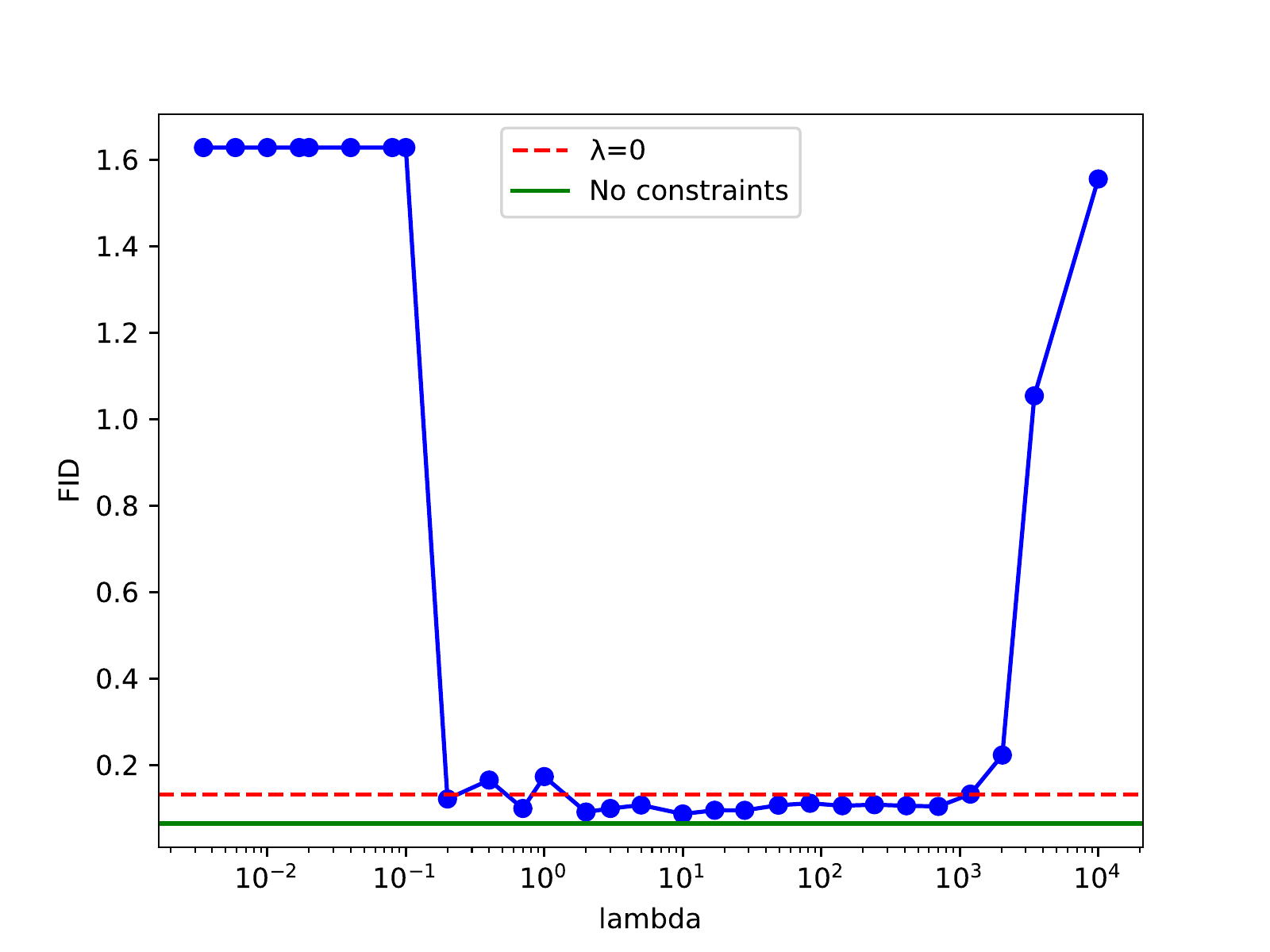}\includegraphics[trim=0 0 0 40, clip,scale=0.36]{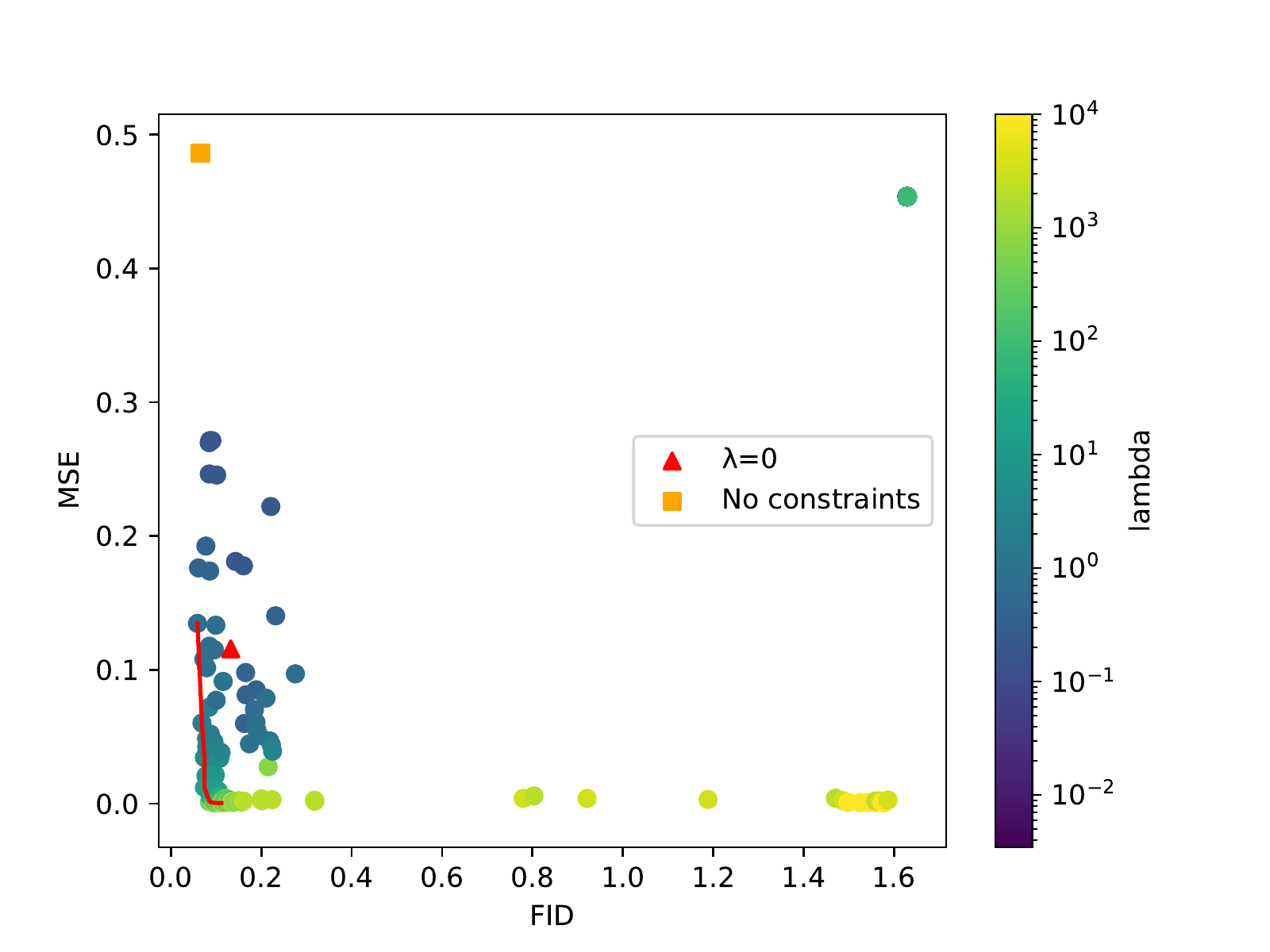}
            \includegraphics[trim=0 0 0 40, clip,scale=0.36]{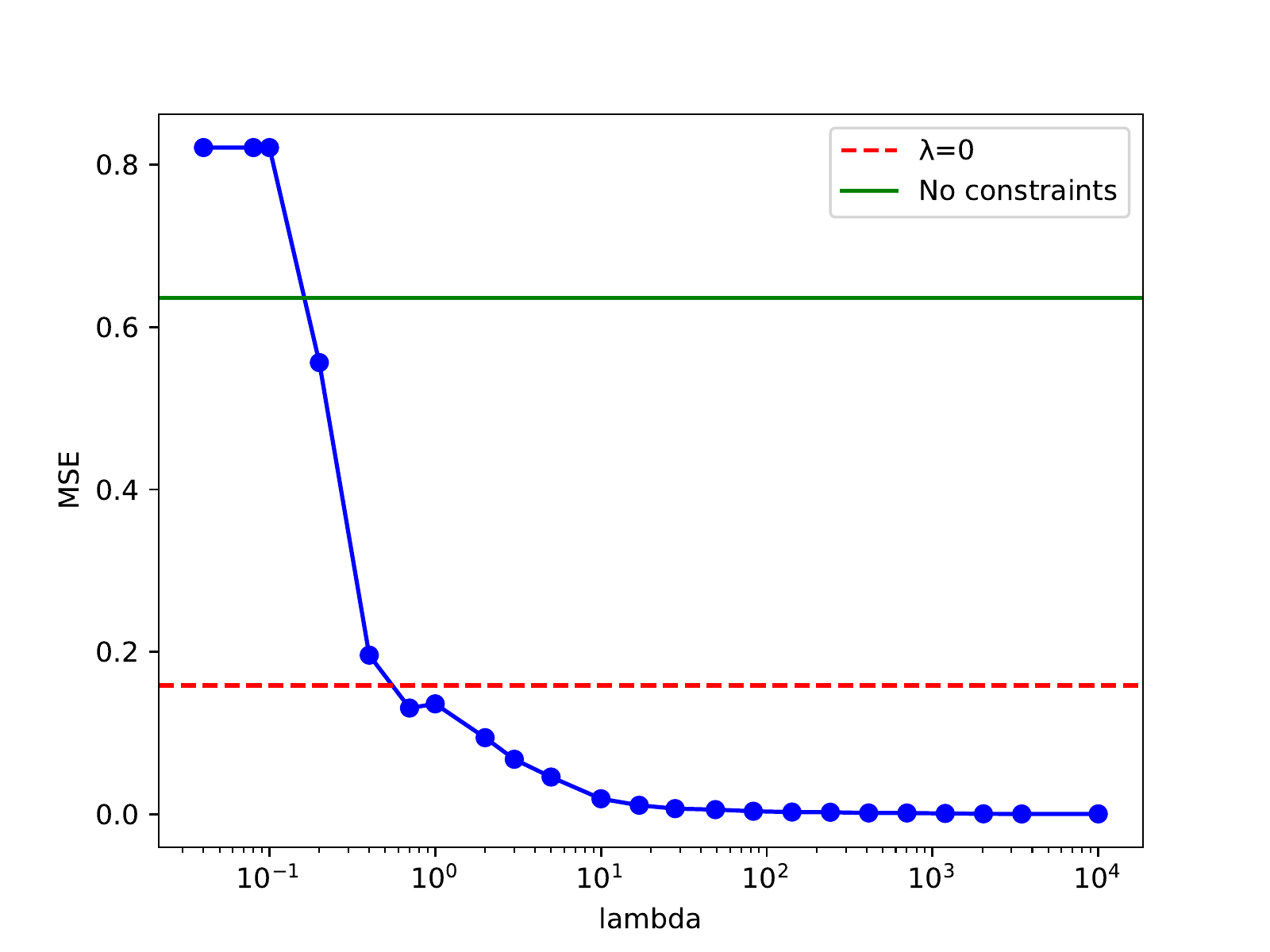}\includegraphics[trim=0 0 0 40, clip,scale=0.36]{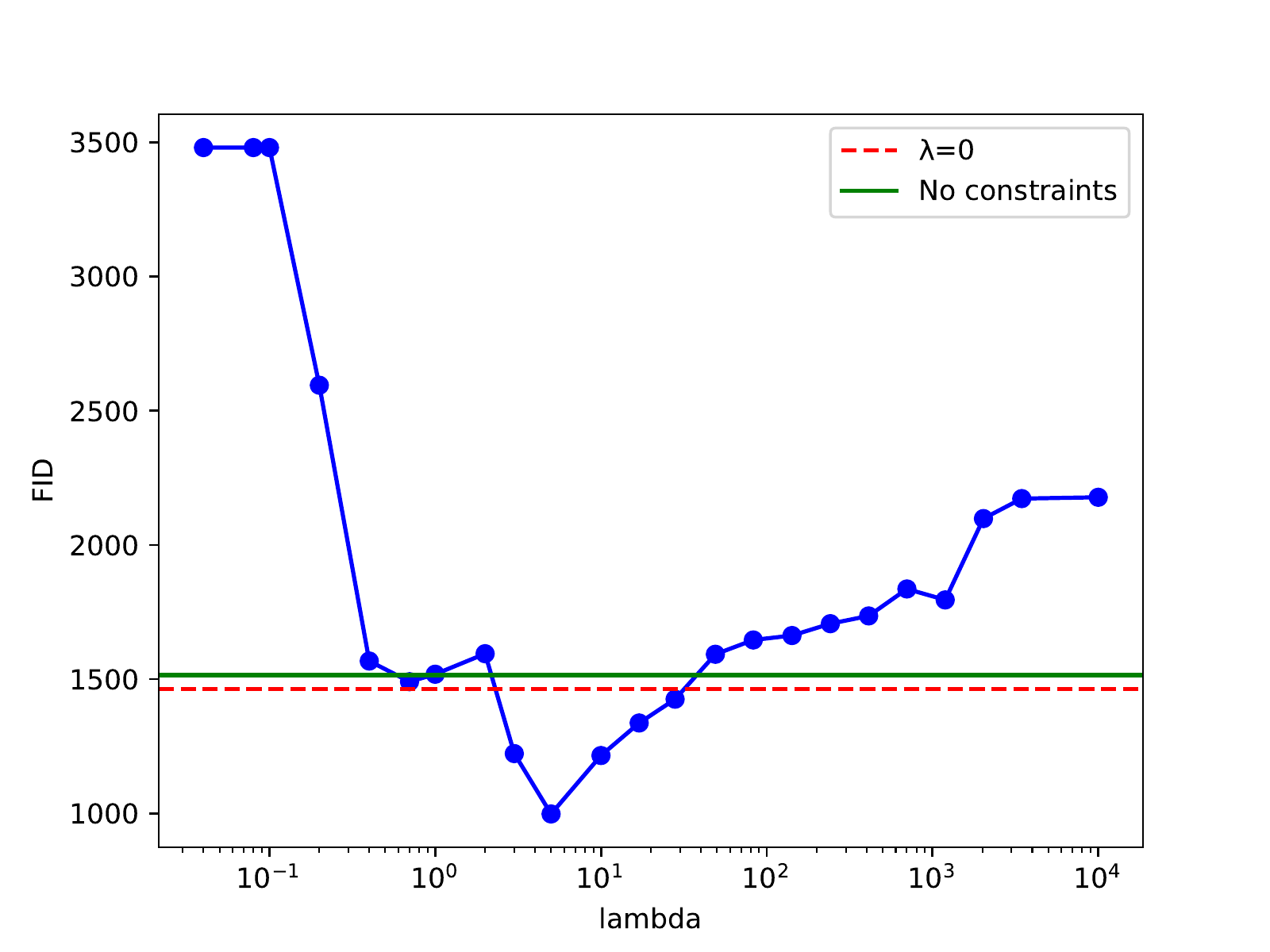}\includegraphics[trim=0 0 0 40, clip,scale=0.36]{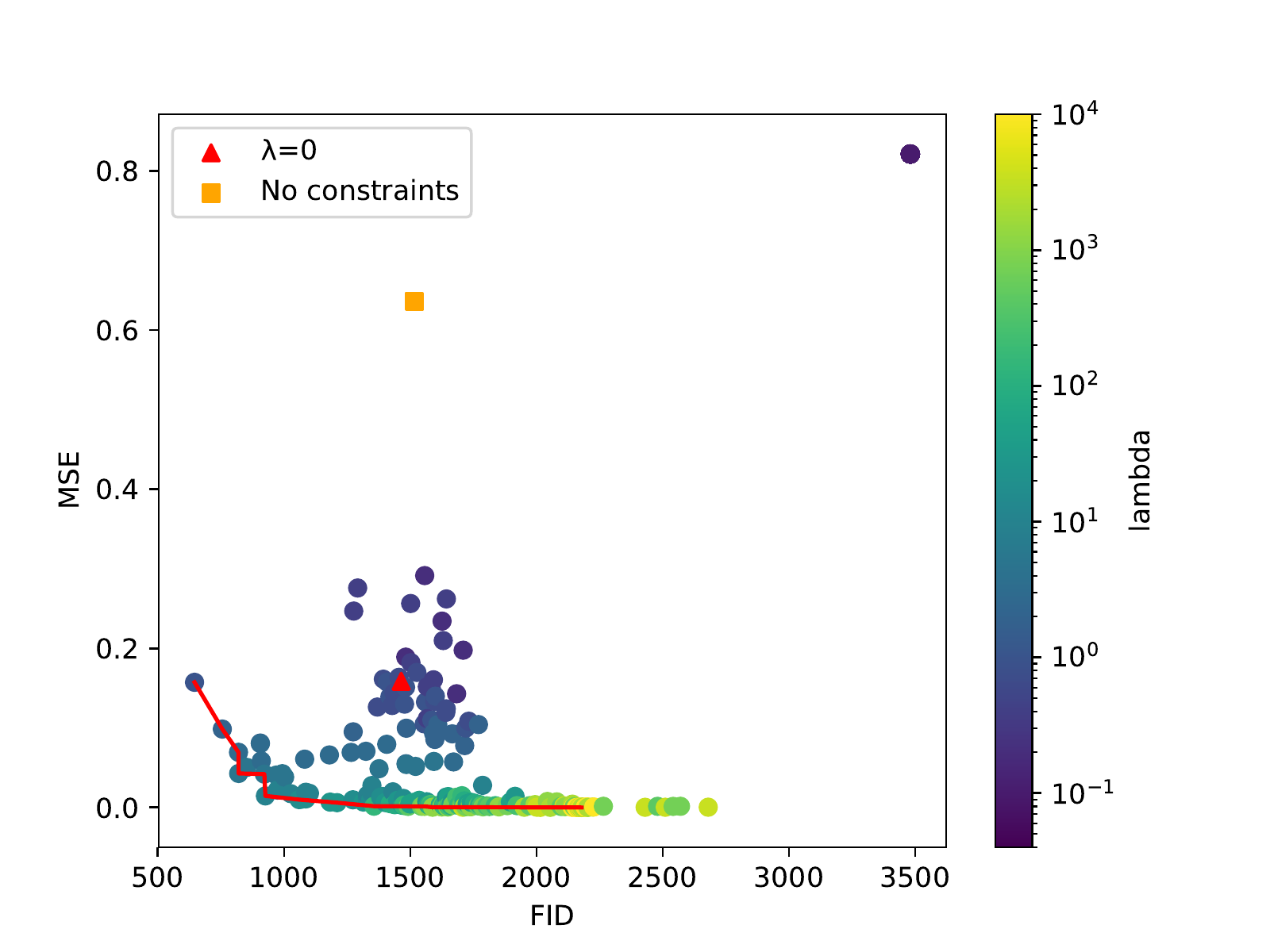}
            \centering
            \caption{MSE (left) and  FID (center) w.r.t. the regularization parameter $\lambda$; MSE w.r.t the FID (right).
            Dataset MNIST (top), Fashion MNIST (bottom).
            %The different orders of magnitude for the Y-axis of the FID is due to the different classifiers used to compute this distances.
            }
            \label{fig:fids}
            \label{fig:mses}
            \label{fig:paretos}
        \end{figure*}

\section{Experiments}

    %\subsection{Experimental setup}

        We experiment on the MNIST \cite{Lecun1998} and FashionMNIST \cite{Xiao2017} datasets, which consist of images of size $28\times 28$px.
        We split the official training set into a new training set (90\%) and a validation set (10\%). The official test set remains our test set. A fifth of each so defined set is used to generated the matrix of constraints $C$ by randomly selecting $0.5\%$ of the pixels. These images are then removed from the training sets, to avoid correlation between real example presented to the discriminator and constrained maps given to the generator.
        
        A discriminator such as presented in DCGAN  \cite{Radford2015} has been chosen with only two convolutional layers of 64 and 128 filters, Leaky ReLU activations and batch normalization \cite{Ioffe2015}. For the generator we retain the DCGAN architecture with a fully-connected layer
        %of $7\times 7\times 7$ units
        and two transposed convolutional layers of 128 and 64 filters with ReLU activations and batch normalization.
        An example of a generated image with the corresponding constraints can be seen in figures~\ref{fig:pixelwise} and~\ref{fig:generated}.
    
    %\subsection{Evaluation}
    %\bigskip
    
        We evaluate our models based on both the satisfaction of the constraints and the visual quality of the generated samples.
        On one hand, we use the mean squared error between the provided constrained values and the constrained pixels in the generated image.
        On the other hand, evaluating the visual quality of an image is not a trivial task \cite{theis2015}. However, the recently developed metric referred to as Fréchet Inception Distance (FID) \cite{heusel2017} seems to be a good metric of performance.
        %and Inception Score\cite{Salimans2016}, have been used to evaluate the performance of generative models. We choose to use the FID since the Inception Score can be a less reliable metric \cite{Barratt}.
%
        %Since the FID require a pre-trained classifier which is not adapted to the databases we used, we trained a simple convnet on these databases and used it to compute a variant of the Fréchet Inception Distance, in which we swap the pre-trained Inception v3 \cite{Szegedy2016} classifier with our network.
        Since using the FID requires a pre-trained classifier,
         %which is not adapted to the databases we used,
        we trained a simple convnet with MNIST/FashionMNIST labels as target. Lower layers of the classifier are then used to produce high-level features needed by the distance:
        %on these databases and used it to compute a variant of the Fréchet Inception Distance, in which we swap the pre-trained Inception v3 \cite{Szegedy2016} classifier with our network.

        \begin{equation}
            \label{eq:fid}
            FID = ||\mu_r - \mu_g||2+Tr(\Sigma_r+\Sigma_g - 2(\Sigma_r\Sigma_g)^{1/2}),
        \end{equation}

        \noindent where $\mu_r$, $\Sigma_r$, $\mu_g$ and $\Sigma_g$ are the mean and the covariance matrices of extracted features obtained on respectively the real and the generated data.
        To overcome classical GANs instability, the networks are trained 10 times and the median of the best scores on the test set at the best epoch are recorded. The epoch that minimizes $\sqrt{FID^2 + MSE}$ on the validation set is considered as the best epoch. %where $FID$ and $MSE$ are the normalized FIDs and MSEs obtained on the validation set.

   % \subsection{Results}
  % \bigskip
    
        Empirical evidences show that with a good choice of $\lambda$, the regularization term helps the generator to learn enforcing the constraints (Fig.\ref{fig:mses}), leading to smaller MSEs than when using the CGAN approach only ($\lambda=0$) and with minor detrimental effects on the quality of the samples (Fig.\ref{fig:fids}). For Fashion MNIST, the regularization term even leads to a better image quality compared to the quality provided by GAN and CGAN approaches.
        Fig. \ref{fig:paretos} illustrates that the trade-off between image quality and the satisfaction of the constraints can be controlled by appropriately setting the value of $\lambda$. Nevertheless, for small values of $\lambda$ the GAN fails to learn and only generates completely black samples. This leads to the plateaus seen for both the MSE and the FID in Fig. \ref{fig:mses}.

\section{Conclusion}

    In this paper, we investigate the effectiveness of adding a regularization term to the conditioning of GANs to deal with cases where only a small subset of the image one wants to generate is known beforehand. Empirical evidences illustrate that the proposed framework helps obtaining good image quality while best fulfilling the constraints compared to classical GAN approaches. In future work, we plan to extend this study to GAN conditioning in situations where no trivial mapping exists between the conditions and the generated samples, such as class-wise conditioning or more structured conditions.

\FloatBarrier
% ****************************************************************************
% BIBLIOGRAPHY AREA
% ****************************************************************************

\begin{footnotesize}

\bibliographystyle{unsrt}
\bibliography{main}

\begin{thebibliography}{10}

\bibitem{laloy2018}
Eric Laloy, Romain H{\'e}rault, Diederik Jacques, and Niklas Linde.
\newblock Training-image based geostatistical inversion using a spatial
  generative adversarial neural network.
\newblock {\em Water Resources Research}, 54(1):381--406, 2018.

\bibitem{mirza2014}
Mehdi Mirza and Simon Osindero.
\newblock Conditional generative adversarial nets.
\newblock {\em arXiv preprint arXiv:1411.1784}, 2014.

\bibitem{Lecun1998}
Yann LeCun, L{\'e}on Bottou, Yoshua Bengio, and Patrick Haffner.
\newblock Gradient-based learning applied to document recognition.
\newblock {\em Proceedings of the IEEE}, 86(11):2278--2324, 1998.

\bibitem{Xiao2017}
Han Xiao, Kashif Rasul, and Roland Vollgraf.
\newblock Fashion-mnist: a novel image dataset for benchmarking machine
  learning algorithms.
\newblock {\em arXiv preprint arXiv:1708.07747}, 2017.

\bibitem{Goodfellow2014}
Ian Goodfellow, Jean Pouget-Abadie, Mehdi Mirza, Bing Xu, David Warde-Farley,
  Sherjil Ozair, Aaron Courville, and Yoshua Bengio.
\newblock Generative adversarial nets.
\newblock In {\em Advances in neural information processing systems}, pages
  2672--2680, 2014.

\bibitem{Yeh2017}
Raymond~A Yeh, Chen Chen, Teck-Yian Lim, Alexander~G Schwing, Mark
  Hasegawa-Johnson, and Minh~N Do.
\newblock Semantic image inpainting with deep generative models.
\newblock In {\em CVPR}, volume~2, page~4, 2017.

\bibitem{Mosser}
Lukas Mosser, Olivier Dubrule, and Martin~J Blunt.
\newblock Conditioning of three-dimensional generative adversarial networks for
  pore and reservoir-scale models.
\newblock {\em arXiv preprint arXiv:1802.05622}, 2018.

\bibitem{Isola2017}
Phillip Isola, Jun-Yan Zhu, Tinghui Zhou, and Alexei~A Efros.
\newblock Image-to-image translation with conditional adversarial networks.
\newblock {\em arXiv preprint arXiv:1611.07004}, 2017.

\bibitem{Yu2018}
Jiahui Yu, Zhe Lin, Jimei Yang, Xiaohui Shen, Xin Lu, and Thomas~S Huang.
\newblock Generative image inpainting with contextual attention.
\newblock {\em arXiv preprint arXiv:1801.07892}, 2018.

\bibitem{Demir2018}
Ugur Demir and Gozde Unal.
\newblock Patch-based image inpainting with generative adversarial networks.
\newblock {\em arXiv preprint arXiv:1803.07422}, 2018.

\bibitem{Radford2015}
Alec Radford, Luke Metz, and Soumith Chintala.
\newblock Unsupervised representation learning with deep convolutional
  generative adversarial networks.
\newblock {\em arXiv preprint arXiv:1511.06434}, 2015.

\bibitem{Ioffe2015}
Sergey Ioffe and Christian Szegedy.
\newblock Batch normalization: Accelerating deep network training by reducing
  internal covariate shift.
\newblock {\em arXiv preprint arXiv:1502.03167}, 2015.

\bibitem{theis2015}
Lucas Theis, A{\"a}ron van~den Oord, and Matthias Bethge.
\newblock A note on the evaluation of generative models.
\newblock {\em arXiv preprint arXiv:1511.01844}, 2015.

\bibitem{heusel2017}
Martin Heusel, Hubert Ramsauer, Thomas Unterthiner, Bernhard Nessler, and Sepp
  Hochreiter.
\newblock Gans trained by a two time-scale update rule converge to a local nash
  equilibrium.
\newblock In {\em Advances in Neural Information Processing Systems}, pages
  6626--6637, 2017.

\end{thebibliography}

\end{footnotesize}

% ****************************************************************************
% END OF BIBLIOGRAPHY AREA
% ****************************************************************************

\end{document}